\documentclass[twoside,11pt]{article}

\usepackage[abbrvbib, preprint]{jmlr2e}
\usepackage[linesnumbered,ruled]{algorithm2e}

\usepackage{tablefootnote}
\usepackage{listings}
\usepackage{hyperref}
\usepackage{graphicx} 
\usepackage{amsmath}
\usepackage[linesnumbered,ruled]{algorithm2e}
\usepackage{booktabs}

\newcommand{\E}{\mathbb{E}}
\renewcommand{\P}{\mathbb{P}}

\begin{document}

\title{Learning from aggregated data with a maximum entropy model}

\author{\name Alexandre Gilotte \email a.gilotte@criteo.com \\
       \addr Criteo\\
       AI Lab\\
       Paris, France
       \AND
       \name Ahmed Ben Yahmed \email ahmed.ben\_yahmed@ens-paris-saclay.fr \\
       \addr Criteo\\
       AI Lab\\
       France
        \AND
       \name David Rohde \email d.rohde@criteo.com \\
       \addr Criteo\\
       AI Lab\\
       France       
       }
       
\editor{}

\maketitle

\begin{abstract}

Aggregating a dataset, then injecting some noise, is a simple and common way to release differentially private data.
However, aggregated data -even without noise- is not an appropriate input for machine learning classifiers.
In this work, we show how a new model, similar to a logistic regression, may be learned from aggregated data only by approximating the unobserved feature distribution with a maximum entropy hypothesis. The resulting model is a Markov Random Field (MRF), and we detail how to apply, modify and scale a MRF training algorithm to our setting. Finally we present empirical evidence on several public datasets that the model learned this way can achieve performances comparable to those of a logistic model trained with the full unaggregated data.

\end{abstract}


\section{Introduction}
\label{sec:Introduction}

\subsection{Learning from aggregated data}

A recent societal trend  is to provide user with more privacy protection, and this leads to rethinking how user data are collected and shared to offer meaningful and provable privacy guarantees. One popular and simple way to provide these guarantees is to aggregate the dataset, discarding the individual records early in the data processing. 
 The resulting aggregated dataset is made of a list of contingency tables, counting examples of the original record-level dataset along several projections. As an example, Table \ref{toyexamplenotagg} shows a toy dataset made of individual records, and  Table \ref{toyexample} contains the "aggregated data" computed from this toy dataset by projecting on each pair of features\footnote{A good reason to aggregate the data is that it is easy to make aggregated data differentially private, by adding some  Laplace or Gaussian noise \citep{dwork2014algorithmic} . The resulting noisy aggregated data may then be shared or published without damaging the privacy of the users.
}.


\begin{table}[hb!]
\centering
\caption{Toy example dataset}
\label{toyexamplenotagg}
\begin{tabular}{*5c}
\toprule{}
 & Feature 1 &  Feature 2   & Feature 3  & label \\
\midrule
Example 1 & "1"  & B   &  a &  1 \\
Example 2 & "2"  & A   &  b &  1 \\
Example 3 & "1"  & B   &  b &  0 \\
Example 4 & "2"  & B   &  a &  1 \\
Example 5 & "1"  & A   &  b &  0 \\
\bottomrule
\end{tabular}
\end{table}

\begin{table}
\centering
\caption{Example of aggregated data}
\label{toyexample}
\begin{tabular}{*4c}
\toprule{}

Feature 1 &  Feature 2   & Examples count  & Sum(label) \\
  "1" & A &  1  & 0 \\
  "1" & B &  1  & 1   \\
  "2" & A &  1  & 1  \\
  "2" & B &  2  & 1  \\
\midrule

Feature 1 &  Feature 3   & Examples count  & Sum(label) \\
"1"   &  a &   1   & 1 \\
"1"   &  b &   2   & 0   \\
"2"   &  a  &  1   & 0  \\
"2"   &  b  &  1   & 1  \\
\midrule
Feature 2 &  Feature 3   & Examples count  & Sum(label) \\
A   &  b  &  2   & 1   \\
B   &  a  &  2   & 2  \\
B   &  b  &  1   & 0  \\
\bottomrule
\end{tabular}
\end{table}

There is however a strong limitation to the usefulness of aggregated data: The vast majority of high performance machine learning algorithms from logistic regression to deep learning require access to disaggregate data (Table \ref{toyexamplenotagg}.) and cannot be trained from aggregate data (Table \ref{toyexample}.).

One notable exception is the ``Naive Bayes classifier'' \citep{lewis1998naive} , which requires only aggregated data. Specifically, it requires, for each feature, one contingency table aggregating on this feature and the label. Naive Bayes heavily relies on an independence assumption between the features of the dataset, and is known for often providing poor quality models on real world datasets.
 We notice also that the aggregated data of Table \ref{toyexample} allows to learn about the correlations between pairs of features, and this information is lost in the Naive Bayes model.

To summarize briefly this method, it consists on retrieving the distribution of maximal entropy among these which would in expectation produce the observed aggregated data; and on predicting with the conditional distribution of the label inferred from this distribution.
This method may be understood as a generalization of Naive Bayes, but takes into account all the observed correlations between features.
 The quality of the obtained model depends on the set of available contingency tables, and we focused in our experiments in the case when there is one contingency table for each pair of feature, as in Table \ref{toyexample}.
 The case when aggregated data are noisy was kept outside of the scope of this paper.


\paragraph{ Paper outline }
The problem and notations are defined in section \ref{sec:Learningfromaggdata}. We then detail our solution in \ref{sec:jointdistribmodel} , discuss its limitations and how to scale it in section \ref{sec:discussion}, and present our experimental results in section \ref{sec:results}. Finally section \ref{sec:related} reviews some related works. 

\subsection{Online advertising, Chrome Privacy Sandbox the Criteo-AdKdd challenge}

Our main motivation in the development of this method comes from the changes happening in the online advertising with the introduction of Google Chrome Privacy Sandbox \citep{privacy-sandbox}.

The online advertising industry currently heavily relies on user data to train machine learning (ML) models predicting the probability that an ad is clicked, as a function of the ad's features. These models are used to price an ad display opportunity, and to bid in real time on these opportunities. 
The Privacy Sandbox would however restrict how advertisers can access to these user data. 
In particular in the FLEDGE \citep{turtledove} proposal from Chrome, the advertisers would still be able to compute their bids as a function of many of the features they use today, but they would not be able to collect an individual record training set 
\footnote{This would be achieved by letting the advertiser upload their bidding model on the browser, where user data would be stored. This model would be applied used by the browser itself to compute a bid, and the advertiser would thus never observe directly the inputs of the model.}. 
Instead, Chrome would provide an "aggregation API", from which advertisers would recover noisy aggregated data similar to these of Table \ref{toyexample}. But this means that new algorithms able to learn directly from these aggregated data would be required.

\paragraph{Criteo Ad-Kdd challenge}

Because learning a bidding model in this setting is a serious challenge for the advertising industry, Criteo proposed in 2021 a public competition where the goal was to learn a model from aggregated data \citep{diemert2022lessons}.
We try to tackle a similar problem as the one from the competition, except that in the AdKKD's competition, a small amount of disaggregate examples were also provided and largely used by the top-performing solutions. Here we focus on a solution which does not require any disaggregate sample.


\section{Problem formalization}
\label{sec:Learningfromaggdata}

Let $X$ a feature vector made of $D$ categorical features, each with $M$ modalities: $ X \in \mathcal{X} := \{1...M\}^D $.  Let $Y \in \{0,1\}$ a binary label, and $\pi$ an unknown joined distribution on $X, Y$.
Let  $(x_i,y_i)_{ i \in 1 \dots n }$  a dataset of $n$ iid samples of this distribution.

From this dataset is computed a list of contingency tables, counting the displays and labels projected on subsets of the feature vector, as in the example of Table \ref{toyexample}.
Formally, we may define these tables as follow:  let $\boldsymbol{\mathcal{\phi}}$ be a finite set of $K$ ``projections'' $\phi_k : \mathcal{X} \longrightarrow\{0,1\} \; for  \; k \in [1..K]$. 
We may think of each function $\phi_k$  as one row in a contingency table: an example $x$ is either counted ( when $\phi_k(x)=1$ ) or not ( when $\phi_k(x)=0$ ) in this row. 
 We then define, for $x\in\mathcal{X}$, the binary vector:
$$ \mathcal{\phi}(x) := ( \phi_k(x) )_ { \{\phi_k\in \boldsymbol{\mathcal\phi} \}} \in \{0,1\}^K .$$

\noindent
We also define the aggregated data as the two vectors\footnote{In the vocabulary of online advertising, $d$ are the counts of \emph{Displays}  and $c$ are the counts of \emph{Clicks} }: 
\begin{align*}
    d := \sum_{i} \mathcal{\phi}(x_i), \\
    c := \sum_{i} \mathcal{\phi}(x_i) \cdot y_i.
\end{align*}

Each coordinate of $d$ (respectively $c$) is thus the count of examples (respectively the examples with label $1$) on one row of a contingency table.  To keep notations compact, we will also note $a$ the concatenation of vectors $c$ and $d$ i.e. $a =\begin{bmatrix}
           c \\
           d \\
         \end{bmatrix}$. 

The vector $\mathcal{\phi}(x)$ thus encodes the list of rows where example $x$ is counted.
In practice, we first choose a list of contingency tables, and define $\phi$ from these tables.

\paragraph{Problem statement}

Our goal is to learn a model predicting the label $Y$ as a function of the features $X$, when \emph{the observed data consist only on the vectors $c$ and $d$ of aggregated data.} 
and the "granular" training set $(x_i,y_i)$ from which it was computed cannot be accessed.
Note that the state space $\mathcal{X}$ and the encoding $\phi$ are assumed to be known, i.e. we know how the data were aggregated.
In other words, and noting $A$ the random variable associated with the observed aggregated data $a$, the only information available at learning time is that we observed the event $ (A =a ) $.

\paragraph{Aggregation on pairs of features}

Of course, the quality of the learned model depends greatly on the list of available contingency tables. 
For example, in the extreme case when we would have one table aggregating on all features, we could rebuild the full original dataset, and could then apply classical ML methods.
At the other extreme, if we only have one table for each feature, aggregating the examples and labels on this feature only, and no other information on the correlation between features, there is little opportunities to learn a meaningful model other than Naive Bayes or some simple ensemble of predictors using one single feature each.

We are thus interested in the intermediate cases, when the tables provide some information on the correlations between features of the $X$ vector, but the full dataset cannot be reconstructed. In practice, we focus in the case when there is one contingency table counting examples and labels for each \emph{pair} of features of $X$.

\begin{table}
\centering
\begin{tabular}{|c|c|}
\hline
Symbol & Meaning \\
\hline
    $\mathcal{X} $ & The space state of feature vectors. \\
    $\mathcal{Y} \equiv \{0,1\} $ & The set of values of the labels\\
    $n$ & The number of samples of the dataset\\
    $\pi$ & The unknown distribution on $\mathcal{X} \times \mathcal{Y}$\\
    $( x_i ,y_i )_{i \in 1...N}$ & Dataset of independent samples of $\pi$\\
    $\boldsymbol{\mathcal{\phi}}$ & Finite family of $K$ binary functions \\ & $\phi_k : \mathcal{X} \longrightarrow \{ 0;1\}$\\
     $\mathcal{\phi}(x)$ & The $K$ dimensional binary \\
     & vector  $( \phi_{k}(x) )_{k \in [1..K]} $\\
     $c$ and $d$ & The observed vectors of aggregated counts\\
     & and labels on projections $\mathcal{\phi}$ \\
     $a$ & the concatenation of vectors $c$ and $d$\\
     $C$, $D$, $A$ & The random variables associated with $c$, $d$, $a$\\
\hline
\end{tabular}
\caption{Notation}
\label{notation}
\end{table}

\section{ Modelling the joint distribution of features and labels }
\label{sec:jointdistribmodel}

Recall that classical ML methods, which directly model the conditional distribution $\P(Y|X=x)$, require observing individual samples $(x,y)$ to define a loss and are thus not applicable here.

Instead, we  will directly model the joined distribution on $X,Y$. Since the aggregated data $A$ are random variables computed from iid samples of $X,Y$, this model is able to fully define the distribution on the random variables $A$, and may be fitted by trying to maximise the likelihood of the observed event $A=a$.

A joined model may also be used to infer the label on a sample $x$, using the Bayes rule to retrieve the conditional $\P(Y|X=x)$ from the joined model.



\subsection{ Log linear model }

The class of models we propose to use is the class of log-linears models whose sufficient statistics are exactly the aggregated data.

Formally, let $\mu,\theta \in \mathcal{R}^K$ two vectors of parameters, and $\pi_{\mu, \theta}$ the parametric distribution on $X,Y$ defined as follow:
\begin{equation}
\label{eq:mu}
 \pi_{\mu, \theta}(x,y) := \frac{1}{ Z_{\mu, \theta} }  \exp \big( \mathcal{\phi}(x) \cdot ( \mu + y \cdot \theta )   \big), 
\end{equation}

\noindent
where $ Z_{\mu, \theta} $ is the normalization constant:
\begin{equation}
\label{eq:normalization}
 Z_{\mu, \theta} \equiv  \sum\limits_{x',y'\in  \mathcal{X}\times\mathcal{Y}} \exp( \mathcal{\phi}(x') \cdot ( \mu + y' \cdot \theta )  ). 
\end{equation}

\begin{remark} This normalisation constant $ Z_{\mu, \theta} $ is a sum on a number of terms exponentially large in the number of features. It is not reasonable to compute it explicitly, except in a small "toy" problems. But as we will see, this is not necessary.
\end{remark}

Models such as Equation \ref{eq:mu} belong to the class of ''Random Markov Fields`` (MRFs).

\paragraph{Motivations for choosing this parametric model}
Let us summarize quickly the reasons why we choose this specific family of parametric models.
\begin{itemize}
    \item The aggregated data are sufficient statistics for this model, making the optimization problem reasonably tractable. In particular, the objective\footnote{To be more precise, this is the case if we allow the parameters to go to infinity, or if there is no 0 cell with a count of 0 in the contingency tables. Also note that the model is slightly over-parametrised: the distribution is unique, not the optimal parameters. In practice we use a regularization, which makes the loss strictly convex and avoids these complications. } is convex, with a uniquely defined optimal distribution $\pi^*$ . 
    \item This model is just rich enough to fit well the available data: at the optimum, the expectation of the random variable "aggregated data" under the model distribution exactly matches the observed data, i.e. $\E_{\pi^*}(A =a)$ \footnote{This is a direct consequence of lemma \ref{lemma:grad}}, and the optimal distribution is the only one in this class of model verifying this equality.
    \item The optimal distribution is also the maximum entropy distribution, among the distributions $q$ verifying the previous property $E_q(A)=a$.
\end{itemize}

\subsection{Conditional distribution of Y knowing X }

As mentioned earlier, Bayes rule may be applied to retrieve the conditional distribution from a joined model:

\begin{align}
\pi_{\mu,\theta}(Y=1|X=x) &= \frac{  \pi_{\mu,\theta}(Y=1,X=x) }{ \pi_{\mu,\theta}(Y=0,X=x) + \pi_{\mu,\theta}(Y=1,X=x) } \nonumber \\
 &= \sigma( \mathcal{\phi}(x) \cdot \theta) \label{eq:predictwithmu}
\end{align}

Here $\sigma$ is the logistic function. We recognize here the shape of a logistic model, on features  $\mathcal{\phi}(x)$. In the case when there is one contingency table for each pair of features, this is exactly a logistic model with a quadratic kernel, which is commonly used to model feature interactions, and is known as a very strong baseline on online advertising datasets\citep{chapelle2014simple}.
The ``only'' difference between our case and a logistic regression is that we also have to fit the distribution on $X$ which we did not observe directly. This additional modelization certainly has a cost, and we thus expect our model to perform less well than this logistic regression. In the experiments of section \ref{sec:results}, we compared the test performances of our method to these of a logistic regression with the same shape.

\subsection{Maximizing the likelihood of the data}
We have seen that a joined model on $X,Y$ assigns a probability $\pi_{\mu,\theta}(A=a)$ to the observed event $A=a$. We may thus define our training loss as the negative log-likelihood of this event:  \footnote{ To be perfectly exact, this argmin in Equation \ref{eq:max_llh_observed} may be either a set of empty. But in practice, we add some L2 regularization, which makes the optimization problem strongly convex and the argmin well defined. }

\begin{equation}
   \label{eq:max_llh_observed} \mu^*,\theta^* :=
   {\rm argmin}_{\mu,\theta} \,  - \log \, \pi_{\mu,\theta} ( A = a)
\end{equation}

While we do not know how this problem could be solved for arbitrary classes of models, for the parametric model of Equation \ref{eq:mu} the gradients of log-likelihood have a close formula:

\begin{lemma}[Gradient of the log-likelihood]
\label{lemma:grad}

$$- \nabla_\mu  \,  \log \, \pi_{\mu,\theta} ( A=a) =  \E_{\mu ,\theta }(D) - d $$  
$$ - \nabla_\theta  \, \log \, \pi_{\mu,\theta} ( A=a) =  \E_{\mu ,\theta }(C) - c $$  

here $\E_{\mu,\theta }(D)$ and $\E_{\mu,\theta }(C)$ are the expectation of the aggregated vectors when the $n$ samples $X_i,Y_i$ come from the model $\pi_{\mu,\theta}$.
\end{lemma}.

This lemma is a direct consequence of the fact that Equation \ref{eq:mu} defines an exponential family \citep{koller2009probabilistic} whose sufficient statistics are exactly the aggregated data, and this is our main motivation for choosing this class of models.

\paragraph{The gradient is the same in the fully observed case }

Another meaningful consequence of the sufficient statistics is that the gradient in lemma \ref{lemma:grad} is \emph{exactly the gradient of log-likelihood of the full training set}:

$$ \nabla_\mu  \,  \log \, \pi_{\mu,\theta} ( A=a  ) =  \nabla_\mu  \,  \log \, \pi_{\mu,\theta} ( {Xi = x_i ,Y_i = y_i}_{ \,i \in 1...n}   ),   $$  
$$ \nabla_\theta  \,  \log \, \pi_{\mu,\theta} ( A=a) = \nabla_\theta  \,  \log \, \pi_{\mu,\theta} ( {Xi = x_i ,Y_i = y_i}_{ \,i \in 1...n}   ),  $$  

\noindent
In other words, it is possible to fit this joined model as well as if we observed the whole dataset! \footnote{However if we had the full dataset, it would be preferable to fit directly a model on $\P(Y|X=x)$, instead of jointly modeling the distribution of $X$ as we do. }

\subsection{Training algorithm}

Markov Random  Fields such as Equation \ref{eq:mu} are not straightforward to train because of the intractable normalization, but several algorithms have been proposed to overcome this issue. In particular, the ''Persistent Contrastive Divergence`` (PCD) method \citep{tieleman2008training} may be directly applied to our case, and we used it in our experiments.
This method consists in running a stochastic gradient descent, using a Gibbs sampling algorithm to estimate the gradient.  


Indeed the gradient of lemma \ref{lemma:grad} is the difference between the aggregated data $c$ and $d$, which are directly observed, and the expectation $\E(C)$ and $\E(D)$ of these aggregated data according to the current model. These expectations do not depend on any additional data, but involve an intractable sum on all possible vector $x$, making an exact gradient computation infeasible.
Instead, a Gibbs sampler is used to draw samples of $X$ and $Y$ from the model, and the expectations $\E(C)$ and $\E(D)$ are estimated by Monte Carlo on these samples. However generating accurate Gibbs samples at each iteration of the gradient may still be prohibitively costly. The key idea of \citep{tieleman2008training} consists in reusing the Gibbs samples of previous iteration to limit this cost: At iteration $t$, a Gibbs sampler is initialized with the samples produced at iteration $t-1$; and these samples are updated with one single step of Gibbs sampling. See the pseudo code of the algorithm in appendix \ref{app:trainingalgo}.

\subsection{ Modifications to the MRF model }
\label{sec:optimizing}

 We have seen in the previous section that the model we want to fit is a  Markov Random Field, and may be trained with the well-known   ``Persistent Contrastive Divergence'' algorithm. However we found that a few modifications to this algorithm were beneficial. Our final training algorithm is in appendix \ref{app:trainingalgo}.
 
 The specificity of our problem is that we only care about the quality of the final conditional model $P(Y|X=x)$. Only the parameter $\theta$ appears in this formula.  This means we only care about accurately estimating $\theta$. In contrast the $\mu$ parameter does not appear in this conditional distribution, so it is not an issue if our estimate of $\mu$ is inaccurate or strongly over-fitted to the data, so long as $\theta$ is not.
  
\paragraph{Model Regularization }
\label{sec:regul}

The main change is the choice of the regularisation of the model. In the case of logistic models, it is well known \citep{chapelle2014simple} that a regularization of the parameters may dramatically increase the performances on the validation set when the number of parameters is large and inputs are correlated. It is especially the case when using a second (or larger) order kernel\footnote{i.e. interactions between features.}. Since our model has the same form as a logistic regression, it most likely would also benefit from the same kind of regularization. 

We used L2 regularization in the experiments due to its simplicity, and because it makes the loss strongly convex. However, we noted that a single regularization parameter was not performing well. Instead we penalized differently the components $\mu$ and $\theta$ of the parameters vector. We thus have two distinct parameters $\lambda_\mu$ and $\lambda_\theta$ :  

\begin{equation}
\label{eq:regul}    
 penalty(\mu,\theta) \equiv \lambda_\theta \cdot \theta ^2 + \lambda_\mu \cdot \mu ^2
\end{equation}

We experimented with varying these two parameters (see table \ref{tab:benchlambda2BanK}) and observed that:
\begin{itemize}
    \item Regularization parameter $\lambda_\theta$ should be set to a value roughly similar to what would be used in a "classical" logistic regression. ( A typical value on a dataset with 20 features and  all crossfeatures lies in the range 100 to 1000)
    \item Regularization parameter $\lambda_\mu$ is best kept much smaller. While no regularization at all can lead to numerical issues, a small value (typically 1 or less) is fine, while higher value may decrease the performances.
\end{itemize}

Our intuition here is that $\mu$ ,which only models the distribution of $X$, may be allowed to overfit the data: if we were able to fully overfit the train set and have $P_\mu(X)  = P_{Train}(X) $, then our model would actually become equivalent to a "classical" logistic regression! While we do not have enough information on the distribution of the train set to achieve this, but a lower regularisation on $\mu$ keeps us closer from this ideal case.

\paragraph{Gradient rescaling}

 Looking at lemma \ref{lemma:grad}, we observe that the gradient moves the parameters of interest $\theta$ when the observed label sum $c$ and the expected label sum are different. There are two possible reasons why they could differ: either because the model $\P_{\theta}(Y|X=x)$ is wrong, and the parameters change should then improve this model, or because the model on $X$ is wrong, thus producing an incorrect estimate of $\E( \phi(X) )$. 
We therefore experimented with a ''rescaled gradient`` formula, which updates the coefficients of $\theta$ only where the ratio of positive numbers is off.
Noting $\hat{d}$ (respectively $\hat{c}$) the expectation of $D$ (and $C$) estimated from the Gibbs samples (See appendix \ref{app:trainingalgo} for details), we replaced the gradient on $\theta$ by the following formula:
\begin{equation}
\label{eq:rmfmodifiedgradient}
 PseudoGradient_\theta := c - \frac{d}{\hat{d}} \hat{c},
\end{equation}


\noindent
where the quotient and multiplication are coordinate-wise. 
This "pseudo gradient" is very similar to the formula used in \citep{diemert2022lessons} by the winners of the AdKdd challenge.
Using this formula does not change the optimum of our problem: indeed if the gradient on $\mu$ is $0$, then $d$ and $d'$ are equal, and this pseudo gradient on $\theta$ becomes equal to the true gradient. 
We observed a significant improvement of the speed of convergence of the parameter $\theta$ when using this pseudo gradient, as shown in figure \ref{NLLHmodifandnotmodifgrad}. 

\paragraph{Preconditioning and fast weights}
We reduced the variance of the gradient by marginalizing out $Y$ in our Monte Carlo estimator of $E_{\mu,\theta}(C)$ (i.e. replacing the sampled $Y$ by the expectation on $Y$).
Our implementation also used a preconditioning of the gradient, multiplying the descent direction by the inverse diagonal of the Hessian. (see appendix \ref{app:trainingalgo} )

 Finally, following the idea of \citep{tieleman2009using}, we experimented with using a larger learning rate for $\mu$ than for $\theta$.
 This also significantly reduced the training time on large models, compared to the `best' common step size (see figure \ref{fastweights}). 

\section{ Discussion on the proposed model }
\label{sec:discussion}

\subsection{ Maximum entropy distribution }
\label{sec:maxentropy}

\paragraph{The model is rich enough to ``fit'' the aggregated data}

A direct corollary of lemma \ref{lemma:grad} is that the expectation of the aggregated data under the optimal (unregularized) distribution $\pi_{\mu^*,\theta^*}$ is equal to the observed aggregated data:

\[
\E_{\mu^*,\theta^*}(D) =d 
\]

\[
\E_{\mu^*,\theta^*}(C)=c.
\]

\noindent
i.e. the model is thus able to ``perfectly fit'' all the observations.

However, $\pi_{\mu^*,\theta^*}$ is not the only distribution on $X,Y$ to have this property. So why should we pick this specific distribution (and so the model of Equation \ref{eq:mu}) instead of another one? 
One possible answer here is to apply the \emph{maximal entropy principle} which advises to select the distribution with the maximum entropy, among the distributions compatible with our data.

If we define $\mathcal{S}$ the set of distributions $q$ verifying $\E_q(D)= d$ and $\E_q(C)= c$, then it is known since Boltzmann that the shape of the distribution of maximal entropy in $\mathcal{S}$ is log-linear in the constraints \citep{jaynes2003probability}. In other words it is exactly the distribution $\pi_{\mu^*,\theta^*}$.



\paragraph{ A generalization of Naive Bayes }
It should be noted that in the case when there is only one  contingency table per feature (i.e. for each feature, one query ``select nb examples, nb positives, feature, grouped by feature'' for each feature in $1\dots D$), the model of Equation  \ref{eq:mu} may be factorized, and becomes equivalent to the well known \emph{Naive Bayes} model.
This should not be a surprise: the ``max entropy'' property we noted above is generalization of the conditional independence hypothesis of the Naive Bayes model.

\subsection{ Limitation of the proposed solution }
\label{sec:limitations}

One caveat to our method  is that there is no guarantees on the performances: it may perform poorly if the true distribution of the data is very different from the max entropy distribution, in other words when there are some important correlations between triplets (or more)  features, unobserved in the available pairwise tables\footnote{Arguably this would be a limitation to \emph{any} method observing only the aggregated data, which does not contain enough information to reconstruct the full distribution.}.

\paragraph{Example of distribution where the pairwise model under-performs }

Here is an example of a distribution of 3 binary features $X_1,X_2,X_3$ and a label $Y$ where the pairwise MRF would perform poorly:

Let $X_1$ and $X_2$ two independent Bernoulli variables each of parameter $0.5$, and $X_3 := X_1 \text{ xor } X_2$.
Let $\P(Y=1|X=x) := 0.25 + 0.5\cdot x3 $ 

It is clear that a logistic regression trained with enough samples should be easily able to approximate this distribution on $Y|X$.

However, we may note here that $X_3$ is also a Bernoulli(0.5) variable, and $X_1,X_2,X_3$ are pair-wise independent (i.e. $X_1\perp  X_3$ and, $X_2\perp  X_3$).  The pairwise aggregation tables are thus not  able to distinguish this distribution from the distribution where $X_1,X_2,X_3$ are jointly independent variables.
Actually, it can be checked that the pairwise MRF becomes here equivalent to the Naive Bayes model on features $X_3$ and $U:= X_1 \text{ xor } X_2$ \footnote{By noting that the aggregation tables are compatible with the distribution where $X_3 \perp U | Y$, $X_1 \perp (X_3, U, Y)$ and $X_2 := U xor X_1 $}.  Its outcome predictions (with infinitely many samples) may be computed: $\hat{y} =0.9$ when $x_3 =1$ , and $\hat{y} =0.1$ when  $x_3 =0$, which do not match the true data.

\paragraph{ Robustness on actual datasets }
Being aware of possible data distributions such as the example above, we were not really expecting our algorithm to perform consistently well on real datasets.
However, when experimenting with various datasets, we did not find low dimensional cases where it performed significantly worse than a logistic.

\subsection{ Scaling to large datasets}
\label{sec:scaling}

One important question for the usability of the model we propose is how it scales to large datasets, such as those used in online advertising.

\paragraph{Scaling in number of features}

In all our experiments, we used one contingency table for each pair of features.
Obviously both the size of the aggregated data and the cost of one iteration of training are thus quadratic in the number of features. 
This mean our method would be difficult to apply to datasets with a large number of features. In our experiments, we obtained rather good results on datasets with up to 19 features, working with still larger datasets would likely require some engineering to limit the number of features and was not investigated in this work.


\paragraph{Scaling with the number of samples of the dataset}

Once the aggregated data are computed, and the meta parameters of the algorithm defined, the training time of our algorithm does not depend on the number of samples in the aggregated data.
However, when this number is increased, it may be beneficial to increase the number of Gibbs samples used internally, and we end up having a longer training time on larger datasets.
Indeed, our algorithm finds a fixed point where the observed sufficient statistics are equal to the expected sufficient statistics under the estimated parameters.  It is an important observation that both of these quantities are noisy.  The observed sufficient statistics are  noisy due to the finite data set size.  The estimate of the expected sufficient statistics is also noisy due to the finite number of Gibbs samples. Except for collecting more data it is not possible to decrease the noise in the sufficient statistics, in contrast the number of Gibbs samples is limited only by the computational budget. In practice there is little benefit in drawing many more Gibbs samples than observed data points as the finite data set size dominates the uncertainty in the estimated parameters.
As an example, we used only $10k$ samples on the \emph{adult} and \emph{ Bank marketing } datasets, and $1M$ on the larger \emph{Criteo-AdKDD} dataset. (see section \ref{sec:results} )


\paragraph{Scaling with the number of modalities per feature}

Let $M$ the maximum number of modalities of one feature of $X$.
The aggregation table on a pair of features is of size $O(M^2)$, and so is the number of weights in the associated ``crossfeature'' of the model. When $M$ is large ( for example, in the challenge dataset $M> 10^5$ ).  In this case $O(M^2)$ become a significant scalability issue. 
Several options exists to get round this problem:

\paragraph{Cross features hashing}
One common way to avoid high dimensional data is the use of the ``hashing trick''.\citep{weinberger2009feature}
Formally, this is achieved by choosing the size $H \in \mathbb{N}$ of the hashing space, and a hash projection matrix $\mathcal{H} \in \mathcal{M}_{K,H}( \{ 0,1 \} ) $ containing exactly one randomly placed $1$ in each row.
We then define the hashed encoding of a vector $x$ as $\phi_{\mathcal{H}}(x) := H \cdot \phi(x)$ , and the hashed aggregated data as the aggregation with the hashed encoding: $d_\mathcal{H} := \sum\limits_i \phi_{\mathcal{H}}(x_i) $ and $c_\mathcal{H} := \sum\limits_i y_i \cdot \phi_{\mathcal{H}}(x_i) $. Likewise, the model learning from the hashed aggregated data may be redefined by replacing $\phi(x)$ by $\phi_\mathcal{H}(x)$ in Equation \ref{eq:mu} and \ref{eq:predictwithmu}. We note here that the conditional model $\P(Y=1|X=x) := \sigma( \phi_\mathcal{H}(x) \cdot \mu_s )$ we obtain exactly matches the shape of a classical logistic with the ``hashing trick''. \footnote{We can retrieve the more standard definition in this context by defining $hash(k)$ as the index of the $1$ on the k-th row of the matrix $\mathcal{H}$, it is straightforward to check that our definition of the hashing trick matches the more common formula $\sigma( \sum_k \theta_{ hash(k(x)) })$; the formulation we use allows us to also define the hashed aggregated data. } The hashing trick is a well known compromise between performances and scalability when training a large scale logistic model; our hope was that it would work as well in our setting.

\paragraph{Hashing and high cardinality correlated features }
We obtained good results with the hashing trick on the \emph{Criteo-Attribution} dataset (see section \ref{sec:results} ), but disappointing results on the \emph{Criteo-AdKdd} challenge data.
 After a careful investigation, we realised that this dataset contains some very strongly correlated features with high cardinality. The hashed aggregated data does not contain all the information about these correlations, and we observed a significant degradation in the $P(Y|X)$ model when including these features.  

\paragraph{ Features pre-encoding }
\label{sec:encodings}
Instead of hashing the crossfeatures, another option is to preprocess the data to directly encode each single large cardinality feature to a more reasonable number of modalities $M'$ . By "reasonable" here, we mean that the crossfeatures built on the encoded feature, which are thus of size $M'^2$, should be kept small enough to easily fit in memory, and a typical value could be $ M'= 1000 $.
On the \emph{Criteo-AdKDD} dataset, we thus used target encodings \footnote{i.e., we computed on a held out set the number of occurrences and the average CTR of each modality, and encoded the raw modalities by discretizing together their count of occurrences and their CTR. } which are a simple way to significantly reduce the number of modalities of large features without degrading too much the final performances.

\paragraph{Training time and parallelization}
Training a large scale MRF using the proposed algorithm is slow. In our experiences it is typically around 100 time slower than training a logistic model having access to the full dataset.
The main bottleneck is the Gibbs sampling step we have to perform on each sample between each gradient iteration.
 But fortunately, this is done independently on each Gibbs sample, which allows to easily parallelize this part of the computation. We thus used a pyspark implementation to run our experiments. The results on the largest \emph{Criteo-AdKDD} dataset required a spark session of 200 machines and a few hours to train with $1M$ Gibbs samples for $500$ iterations.

\section{Experiments}
\label{sec:results}

\subsection{Datasets}
We ran our experiments in 4 public datasets:
\begin{itemize}
        \item \emph{ Criteo-Attribution } is an advertising dataset released by Criteo, with examples describing a display and with predicted a "click" label. It contains 16M examples with 11 categorical features.\footnote{\url{https://ailab.criteo.com/criteo-attribution-modeling-bidding-dataset/}}
        \item \emph{ Adult dataset } \citep{Dua:2019} is a dataset used to predict whether a given adult makes more than \$50,000 a year. It contains 48842 examples (train:32561 , test:16281) with 14 features.
        \item \emph{ Bank marketing dataset } \citep{Dua:2019,moro2014data} contains 41188 examples with 20 features describing marketing campaigns of a banking institution, its label is 1 on clients who will subscribe.  
        \item \emph{ Criteo-AdKdd challenge }\citep{diemert2022lessons} is the largest dataset we used. The train set contains $80M$ samples, with 19 features each. 
\end{itemize}

\subsection{Baselines and skylines}

On each dataset, we compared the performances of several models:

\begin{itemize}
\item \emph{MRF} is the model presented in this work, learned only from aggregated data.
\item \emph{NB} is the Naive Bayes baseline
\item \emph{B2f} is a logistic regression using only two features, and an interaction between these features. We trained one such model for each pair of features, and reported the one with the best test score. Note that all these models could be trained directly from the pairwise aggregated data.
\item \emph{Logistic} is a classical logistic regression with exactly the same shape (Equation \ref{eq:predictwithmu}) as the MRF, but trained with the whole non - aggregated dataset. It is expected to perform better than the MRF because it is not constraint to use only the aggregated data.
\end{itemize}

While logistic regression is not the state of the art and might get outperformed by more complicated models, it is still a very solid model, which was still widely used in the advertising industry only a few years ago. Retrieving performances similar to a logistic regression on a large scale aggregated dataset seems both challenging enough, and good enough for practical applications. We thus did not compare to more advanced classical ML methods such as Deep Learning, which would not be applicable to the aggregated dataset anyway.

\paragraph{Reported metric}
We reported the \emph{normalized log-likelihood} of the $P(Y|X=x)$ model, computed on a test set:
$$ NLLH \equiv \frac{ LogLikelihood( Prediction(X), Y )  }{ Entropy(Y) } - 1 $$.  
Note that is it simply an affine transformation of the usual log-likelihood, which we find slightly easier to compare between datasets. A higher score a better.

\paragraph{Preprocessing}
We picked these datasets because they mostly contained categorical features. The few continuous features were discretised by computing deciles, and then treated as categorical.

\paragraph{Preprocessing on the Criteo-AdKdd challenge dataset }
This dataset is available in two versions: the noisy pre-aggregated data which were available to the challengers, and the full un-aggregated which was published with \citep{diemert2022lessons}. 
We found that our method was not performing well with the pre-aggregated data. As explained in section \ref{sec:encodings} , the main issue was the presence of large cardinality, strongly correlated features. Inclusion of these features in the model was degrading its performances, and the best MRF result reported in table \ref{tab:NLLHresults} was obtain by removing these features. The results obtained with the pre-aggregated data and raw features are reported under \emph{Criteo-AdKdd, (raw)}.

With the release of the whole dataset, the problem above can be avoided by computing some target encodings and aggregating on these target encodings. We used $4M$ kept out lines to precompute these encodings, reducing the cardinality of each encoded feature below 1000, aggregated the remaining dataset and trained on the resulting aggregated data.
 We also re-trained the logistic ''skyline`` with the same encoded features. The results of these tests are reported under  \emph{Criteo-AdKdd, (target encoded)}.

 As a side note, we also tried to compute the target encoding directly from the pre-aggregated data released with the challenge, and to re-aggregate these data. It did not work well, for several reasons: first there is an over-fitting issue when the same data are used to compute the encodings and for training. Second, when re-aggregating, we were summing lots of instances of the privacy preserving noise (a Gaussian of std=17) together. While this noise is low enough to have a small impact when using directly the raw features, (see the experiments in \citep{diemert2022lessons}) the variance of the summed noises was considerably larger, and it became an issue. 
 

\paragraph{Reproducibility}
All the datasets we used are public, and 
our code and notebooks are available on \href{https://github.com/criteo-research/ad_click_prediction_from_aggregated_data}{our github}.

\subsection{Results}

Table \ref{tab:NLLHresults} summarizes the test normalised log-likelihood obtained on each dataset. On all datasets, the MRF consistently outperforms the simple baselines. More strikingly, its performances are comparable to the logistic on \emph{Bank marketing} and \emph{Adult}, and only slightly below on \emph{Criteo-Attribution}. The performance gap with the logistic is more noticeable on \emph{Criteo-AdKdd}. We already discussed earlier why it under-performs on \emph{Criteo-AdKdd raw}, and recomputing the aggregation on target encodings brought a fair improvement. We also note that on this dataset the logistic trained with the raw features (line \emph{Criteo-AdKdd raw}) is better than the logistic trained with target-encoded features. While this is not really surprising (the target encodings lose part of the information), it means that the ''skyline`` is lower with these encodings, and finding a better way to define these encodings could therefore be a lead to further improve the MRF.

\begin{table}
\centering
\caption{\label{tab:NLLHresults} NLLH test for different models and several datasets }
\small
\begin{tabular}{ |c||llll|  }
 \hline
 Different & \multicolumn{4}{c|}{NLLH test}\\
datasets &  Logistic & MRF &  Naive Bayes &  B2f \\
 \hline
 Criteo-Attribution  & 0.091 & 0.086  & 0.053  & 0.048 \\
 Bank marketing   & 0.427 & 0.421 & 0.189  &  0.318 \\
 Adult  & 0.421 & 0.416 & 0.213 &  0.297 \\
  Criteo-AdKdd, (raw) & 0.311  & 0.265  & 0.175\tablefootnote{\label{note1}without early stopping it does completely diverge } & 0.139  \\
  Criteo-AdKdd, target encodings   & 0.303  & 0.285 & 0.191\textsuperscript{\ref{note1}} & 0.168   \\

 \hline
\end{tabular}
\end{table}

\paragraph{Regularization on $\mu$ and $\theta$ }
Table \ref{tab:benchlambda2BanK} shows how the change of the two regularisation parameters defined in Equation \ref{eq:regul} impacts the performances on test data, on the \emph{Bank marketing} dataset.
It can be observed that while some regularization on $\theta$ is important to get good results, the regularization on $\mu$ is better kept at a much lower level.

\begin{table}
\centering
\caption{\label{tab:benchlambda2BanK} Effect of parameters $\lambda_{\mu}$ and $\lambda_{\theta}$ on NLLH (\emph{Bank marketing  dataset}) }
\begin{tabular}{c|cccc}
 \hline
 
 \hline
& $\lambda_{\mu} = 1$  & $\lambda_{\mu} = 4$  &$\lambda_{\mu}= 16$  & $\lambda_{\mu} = 64$ \\ 
\hline
$\lambda_{\theta} = 1$   & 0.338 & 0.322 & 0.264 & 0.174  \\
$\lambda_{\theta} = 4$   & 0.405 & 0.405 & 0.389 & 0.337  \\
$\lambda_{\theta} = 16$  & 0.416 & 0.415 & 0.406 & 0.382 \\
$\lambda_{\theta} = 64$  & 0.421 & 0.420 & 0.419 & 0.406  \\
\end{tabular}
\end{table}

%

\paragraph{Performances VS number of Gibbs samples}

\begin{figure}[h]
  \includegraphics[width=1.0\linewidth]{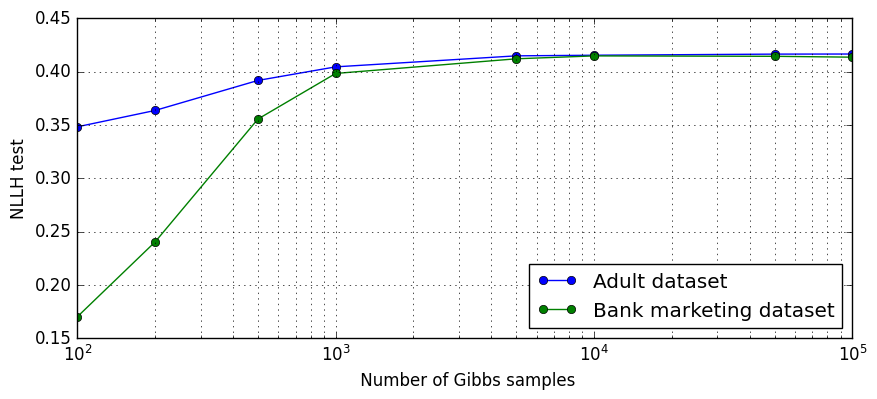}
  \caption{Effect of number of Gibbs samples }
  \label{NLLHvsnbsamples}
\end{figure}

Figure \ref{NLLHvsnbsamples} shows the test performance on datasets: \emph{Bank marketing } and \emph{adult} as a function of number of Gibbs samples used to estimate the expected gradient. 

\begin{figure}[h]
  \includegraphics[width=0.95\linewidth]{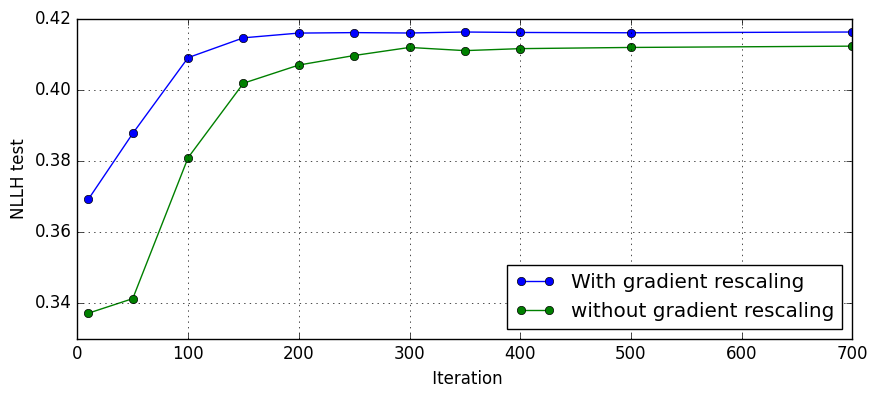}
  \caption{Effect of the gradient rescaling on the Adult dataset}
  \label{NLLHmodifandnotmodifgrad}
\end{figure}

\paragraph{Usefulness of the rescaling trick}
Figure \ref{NLLHmodifandnotmodifgrad} shows the test performance on the Adult dataset as a function of number of training iteration, with either the correct gradient of Equation \ref{lemma:grad} or the ''rescaled`` gradient of Equation \ref{eq:rmfmodifiedgradient}. Clearly the ''rescaled`` gradient is converging faster in this case, which is typical of what we observed.

\begin{figure}[h]
  \includegraphics[width=0.95\linewidth]{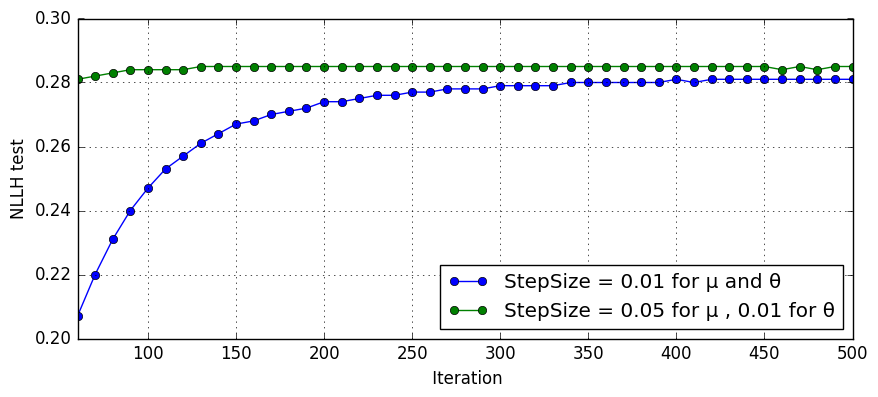}
  \caption{Effect of using different step sizes on Criteo-AdKdd (with target encodings) dataset}
  \label{fastweights}
\end{figure}

\paragraph{Different step sizes on $\mu$ and $\theta$}
Figure \ref{fastweights} shows that a higher stepsize on $\mu$ than on $\theta$ was highly beneficial on the challenge dataset. It should be noted that increasing both step sizes to the $0.05$ value used on $\mu$ lead the model to diverge.

\begin{figure}[!ht]
    \includegraphics{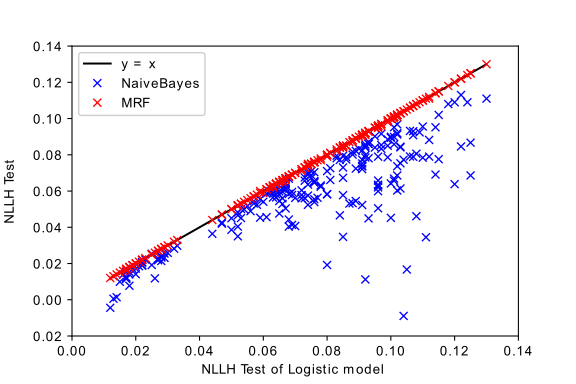}
  \caption{Comparing MRF and logistic on 220 triplets of features }
  \label{all_triplets}    
\end{figure}

\paragraph{ Searching for instances of distributions where the MRF fails }
We noted in section \ref{sec:limitations} that the MRF may under-perform on some distributions with only 3 features and a label. We wonder if we would find such distributions in our data. On the dataset of the AdKdd challenge, we selected the 12 features with less than 100 modalities, and tested each triplet of these features. For each triplet, we trained a logistic and a MRF with these three features, and compared the performances on the test set. We also trained a Naive Bayes with the same features.
 These results are displayed on figure \ref{all_triplets}: the MRF was able to almost match the logistic on all 220 tested triplets (The maximum difference was smaller than 0.001)

\section{Related work}
\label{sec:related}

\paragraph{Differential privacy and learning private models}

Differential privacy \citep{dwork2006calibrating} is a mathematical definition of privacy, which relies on randomizing the publicly shared data to obtain provable users protection. 

Various methods to make a machine learning model differential private have been proposed, including result perturbation, objective perturbation \citep{chaudhuri2011differentially}, and  injecting noise to the gradient during the learning procedure\citep{abadi2016deep}.
One shortcoming of these methods however is that they may be practically difficult to apply if the goal is to let a third party learn a model on the private data. (That would require a mechanism to allow the third party query the noisy gradients of its own model, thus splitting the ML infrastructure between both actors.) 

A much simpler way to release data with differential privacy consists of sharing aggregated data (typically counts as in listing \ref{toyexample}) with some additive noise\citep{dwork2006calibrating}. Adding iid noise from a Laplace distribution provides $(\epsilon,0)$ privacy, while iid Gaussian noise provides $(\epsilon,\delta)$ differential privacy.

\paragraph{ Ecological inference }
Learning about individual-level outcomes from aggregated data has been known historically as the \emph{ecological inference} problem.\citep{king2004ecological,freed} But the proposed methods were typically applied to very low dimensionality problems, and it is unclear how they would scale to  large modern datasets.

\paragraph{Generative and discriminative models} 
A discriminative model approximates the conditional distribution $\P(Y=y|X=x)$ of a label $Y$ as a function of the feature vector $x$. Learning such model is the focus of supervised learning. A generative model instead approximates the joined distribution on $\P(X,Y)$ of the variables of interest. While it is possible to use a generative model to predict $\P(Y=y|X=x)$ with the Bayes rule, these models usually perform less well than discriminative models optimized for this specific task.
 Theoretical properties of the two approaches are compared in \citep{jordan2002discriminative,efron1975efficiency}. 
Hybrid approaches have also been proposed, e.g. \citep{lasserre2006principled}.

\paragraph{Partially observed datasets}
Learning a model in settings where the dataset is not fully observed has been the focus of multiple recent works, such as ``Multiple Instance Learning'' \citep{zhou2004multi} or ``Multiple Instance Regression'' \citep{Ray2001Multi}, or \citep{zhang2020learning}.
Closer from our work, the winners of the Ad-Kdd challenge leverage the fact that a logistic model may be learned using only the aggregated labels and non aggregated \textit{unlabelled} samples \citep{diemert2022lessons}. In \citep{whitaker2021logistic}, the authors study the case when the features $X$ are continuous, and aggregated tables on discretizations of these features are available. The method they propose assumes that the features $X$ follows a multivariate Gaussian distribution, whose parameters are retrieved from the available tables. From this model on $X$ and the aggregations, they can learn a simple logistic model $\sigma(x.\theta)$. 
This however only allows to learn rather simple models, with up to $20$ learnt coefficients in the experiments they present. \footnote{ In contrast, the class of models we use is much richer, with typically thousands to millions of parameters, allowing a much better fit in the large data setting.}



\paragraph{Markov Random Fields}
Markov Random Fields (MRF) \citep{koller2009probabilistic,NotesOnGraphicalModels} are probabilistic models on a set of variables $(X_1,...X_n)$, which may be factorized as a product of parametric positive functions on subsets of the full vector of features$(X_1,...X_n)$.
They naturally appears as maximum entropy distributions under some constraints on the expectation of some random variables, and have thus been long studied. 
While they form an exponential family and have sufficient statistics\citep{koller2009probabilistic}, training a MRF is often not easy, because the exact gradient is usually intractable. Many algorithms have been proposed to avoid this issue, either based on sampling such as Persistent Contrastive Divergence \citep{hinton2002training, tieleman2008training, tieleman2009using,salakhutdinov2009learning}, 
or based on other approximations such as pseudo likelihood or variational methods \citep{ackley1985learning,della1997inducing,mckenna2019graphical,murray2012bayesian,grelaud2009abc}.

\paragraph{Collective Graphical Models}
The idea that aggregated data are the sufficient statistics of a well chosen MRF, which may be learned from these aggregated data only, have already been explored in the ''collective graphical model`` literature \citep{sheldon2011collective,mckenna2019graphical,bernstein2017differentially}. The main difference with our work is that they focus on learning a model on the full set of variables, while we are only interested in learning  correctly the conditional distribution $\P(Y=y|X=x)$ on one specific component $Y$.  



\section{Conclusions}
\label{sec:conclusion}

The empirical results we have obtained on various datasets show that the method proposed in this work can be effective on real data distributions, and is typically able to retrieve the performances of a logistic on small to medium datasets, provided aggregations on all pairs of features are available.
On the large Criteo-AdKdd dataset, there is still a noticeable lag between the performance of our method and these of a logistic regression trained with access to the whole dataset.
Despite these lower performances, we believe that our method would be the most promising way to learn a model in FLEDGE through the aggregation API. 
Whether this is sufficient to sustain a viable targeted ad ecosystem relying only on aggregated data is beyond the scope of this work, as many other factors will also limit further the performances, and learning a model in this context is only one among many technical complications.

Several points have not been addressed here and could be the topic of future work:
first, in all our experiments, we relied on held out unaggregated data to choose the meta-parameters of our method and assess the final quality of the model, and methods or heuristics to evaluate the quality of the model without relying on such data would be needed.
Finally, we did not explore in this article the case of noisy aggregated data, and more work would be needed to efficiently adapt our method to these cases.

\vskip 0.2in

\bibliography{sample}

\appendix

\section{Details on the experiments}
\label{sec:exp-trainingmodels}
\paragraph{Training logistic skylines}
 All logistic models were trained with 200 iterations of L-BFGS. The regularization parameter was benched on a log 2 scale, and we kept only the best model on the validation set.
Our Naive Bayes implementation also includes a L2 regularization that we benched in the same way. \footnote{Our implementation of Naive Bayes was using a L2-regularized logistic regression per for each variable. The parameter was of those logistics was the same, and chosen to get best test performances on the Naive Bayes, not on the individual logistics.}

All logistic models were trained long enough to observe  convergence of their results. The only important meta-parameter was the $L2$ regularisation. We did a grid search for this parameter, and reported only the best result.

\paragraph{Training the MRFs}
MRFs have several important meta-parameters, and choosing them carefully matters. After a bit of experimentation, we applied the following heuristics:
\begin{itemize}
    \item Set the stepsizes to $\frac{1}{K}$, where $K$ is the number of aggregation tables. Alternatively, increase the stepsize on $\mu$ to $5$ time this value. (We obtained the fastest convergence this way, but it might be sometimes instable. )
    
    \item Increase the number of Gibbs samples up to $10\%$ of the number of samples in the true data, if possible. (On the AdKDD challenge, the reported results used ''only`` $1M$ samples. We tried increasing further but it did not seem to improve our results.
    \item Keep the regularization on $\mu$ low. A value of $1$ was enough to prevent numerical instability, and increasing it was only harmful.
    \item Train as long as possible. On the challenge dataset, obtaining the best results required either $250$ iterations while using the increased $\mu$ stepsize,  or running for more than $1000$ iterations (with the default stepsizes above).
    \item We used a grid search for the  regularization on $\theta$. The best value is typically the same magnitude as the best regularization for a logistic.
    
\end{itemize}

\paragraph{Training Naive Bayes}
Noting that our method is a generalization of Naive Bayes, we used the same code and method to train the Naive Bayes models. However, we noted that on most datasets, the performances of Naive Bayes increase at the beginning of the training, quickly reach a maximum and then degrade quickly degrade to worse (in log loss) than a constant model. We thus used early stopping and reported the best test value during the training. (This means that the true performances of a NB model would be in practice \emph{worse} than what we report, unless a very good stopping heuristic is found.)

\section{Proof of lemma \ref{lemma:grad}  }

The main ingredient in this proof is the observation that the gradient of the likelihood of the full un-aggregated dataset $(X_i,Y_i)_{i \in 1 \dots n}$ depends only on the aggregated data $A$.

Let us for now assume that the dataset is fully observed (ie, not aggregated.) We could then compute the log-likelihood of the full dataset according to the model. It is defined as:

$$ \mathcal{L}( \mu,\theta, \{(x_i,y_i)\}_{i\in 1...n}  ) \equiv \sum\limits_{ i \in 1 ...n }  \log\big( \pi_{\mu,\theta} ( x_i , y_i) \big). $$

We would in this case fit the model by solving the following optimization problem:
\begin{equation}
   \label{eq:maxfull}
   \text{Find } \mu^* ,\theta^* \in
    argmin_{\mu,\theta} -\mathcal{L}( \mu,\theta, \{(x_i,y_i)\}_{i\in 1...n}  ).
\end{equation}

\noindent
And we can now compute the gradient of this full likelihood:

\begin{lemma}[Gradient of the full log-likelihood]
\label{lemma:grad-full}
$$ \nabla_\mu - \mathcal{L}( \mu,\theta, (x_i,y_i)_{i\in 1...n} =  \E_{\mu,\theta}(D) - d  $$
$$ \nabla_\theta - \mathcal{L}( \mu,\theta, (x_i,y_i)_{i\in 1...n} =  \E_{\mu,\theta}(C) -c,  $$

\noindent
where $\E_{\mu,\theta}$ is the expectation of random variable when data are sampled according to the model $\pi_{\mu,\theta}$.
\end{lemma}

Proof of this lemma is detailed in the next paragraph. It involves only simple calculus.

Noting that the expectation $\E_{\mu,\theta}(A)$  does not depend on the dataset, \emph{the gradient of the loss depends on the dataset only through the aggregated data $a$.} 

In particular, these gradients are the same for any possible dataset $x_1,y1, ... x_n,y_n$ compatible with the observed aggregations $a$. This means that all the likelihoods (without log) of all those datasets are equal up to some multiplicative factors (which disappear when we apply the log and derive).
Noting that the likelihood of the event $A=a$ is the sum of the likelihood of all those individual compatible datasets, its likelihood is also, up to a multiplicative constant, equal to the likelihood of any compatible individual dataset. As a consequence:
$$ \nabla_\mu \mathcal{L}( \mu,\theta, \{(x_i,y_i)\}_{i\in 1...n}) =  \nabla_\mu Log \pi_{ \mu,\theta} (A=a) $$
$$ \nabla_\theta \mathcal{L}( \mu,\theta, \{(x_i,y_i)\}_{i\in 1...n} )=  \nabla_\theta Log \pi_{ \mu,\theta} (A=a) $$
which concludes the proof.

In other words, we just verified that the aggregated data $a$ form a sufficient statistic for estimating the parameters $\mu,\theta$. This was actually expected because we have an exponential family.
This property means that problems \ref{eq:maxfull} and \ref{eq:max_llh_observed}
are equivalent for this family of distributions, which can be solved by gradient descent as well as when the full dataset is available.

\paragraph{Proof of lemma \ref{lemma:grad-full} }

Let us derive the log-likelihood with respect to  $\theta$:

\begin{align*}
\frac{ \partial \mathcal{L}  }{ \partial  \theta  }  &= \sum\limits_i  \frac{ \partial \log( \pi_{\mu,\theta}  (x_i,y_i) )  }{ \partial  \theta} \\
&= \sum\limits_i \frac{ \partial  \mathcal{\phi}(x_i) \cdot(   \mu + y_i \theta  )  }{ \partial  \theta } - n \times\frac{ \partial \log(Z)  }{ \partial  \theta  } \\
&=   \sum\limits_i y_i \phi(x_i)    - n \times\frac{ \partial \log(Z)  }{ \partial  \theta  } \\
&=   c    - n \times\frac{ \partial \log(Z)  }{ \partial  \theta  } 
\end{align*}

where $Z$ is the normalisation factor $Z = \sum\limits_{x,y} \exp( \phi(x) \cdot(   \mu + y \theta  ) ) $

Noting that:
$$\frac{ \partial Z  }{ \partial  \theta }  = \sum\limits_{x,y} y \phi(x) \exp(  \mathcal{\phi}(x) \cdot (\mu+y \theta  ) )$$
We get:
\begin{align*}
     n \times\frac{ \partial \log (Z)  }{ \partial  \theta  }  &= n \times \sum\limits_{x,y} y \phi(x)  \cdot  \frac  {\exp( \mathcal{\phi}(x) \cdot ( \mu + y \theta )  )}{Z} \\
    &= n \times \sum\limits_{x,y} y \phi(x) \pi_{\mu,\theta}(x,y) \\
    &= \E_{\mu,\theta}( C )
\end{align*} 

And thus $$\frac{ \partial \mathcal{L}  }{ \partial  \theta  } = c - \E_{\mu,\theta}( C )$$

The case of the derivation with respect to  $\mu$ is similar, it is thus left to the reader.

\section{Training algorithm}

Algorithm \ref{algo:train} details the final training algorithm we used, including all the modifications to the original Persistent Contrastive Divergence.

\label{app:trainingalgo}
\begin{algorithm}[h]
\SetKwFor{RepTimes}{repeat}{times}{end}
\SetKwComment{Comment}{//}{}
\SetKwInOut{Input}{Input}
\SetKwInOut{Output}{Output}
 \Input{ initial parameter $\mu$ and $\theta$, \\
         aggregated data $d$ and $c$,  \\
         $n$ number of samples in the aggregated data \\
         $n'$ number of Gibbs samples \\
        initial Gibbs samples $\tilde{x_1} ... \tilde{x_{n'}} $\\
        step sizes $\alpha_\mu$ and $\alpha_\theta$ \\
        regularisation $\lambda_\mu$ and $\lambda_\theta$ \\
        number $T$ of training iterations
}  
 \Output{} 
 \RepTimes{T}{
  \tcp{Gibbs step}
 Draw  $\tilde{y_i} \sim \pi_{\mu,\theta}(Y|X=\tilde{x}_i)$\\
\For{each feature $d$}         {
    Replace $\tilde{x}_i^d $ by a sample of $P_{\mu,\theta}( X_d | X^{-d} = \tilde{x}_i^{-d} , Y= \tilde{y_i} )$  \\
        }
    \tcp{Monte Carlo estimator of $\E(D)$}
    Set $\hat{d} \longleftarrow \sum\limits_i \phi(\tilde{x}_i) $ \\ 
    \tcp{Marginalised Monte Carlo estimator of $\E(C)$}
    Set $\hat{c} \longleftarrow \sum\limits_i \sigma( \phi( \tilde{x}_i ) \cdot \theta ) \cdot \phi(\tilde{x}_i) $ \\ 

    \tcp{Gradient on $\mu$}
    Set $g_\mu \longleftarrow   \dfrac{n}{ n' } \cdot \hat{d} - d + 2\lambda_\mu $  \\ 
    \tcp{Pseudo Gradient on $\theta$ }
    Set $g_\theta \longleftarrow  \dfrac{d}{\hat{d}} \hat{c} - c + 2\lambda_\theta $   \\ 
    \tcp{Gradient preconditioning}
    Set $dir_\mu \longleftarrow  \dfrac{g_\mu}{ \hat{d} +  2\lambda_\mu }$ \\
    Set $dir_\theta \longleftarrow  \dfrac{g_\theta}{ \hat{c} +  2\lambda_\theta }$ \\
    \tcp{gradient step}
    Set $\mu \longleftarrow  \mu - \alpha_\mu dir_\mu  $ \\
    Set $\theta \longleftarrow  \theta - \alpha_\theta dir_\theta  $ \\

}
\caption{Training Algorithm}
\label{algo:train}
\end{algorithm}

\paragraph{ Details on the preconditioning }

We preconditioned the gradient with the inverse of  diagonal Hessian.
This may be computed as follow:

\begin{align*}
    \frac{ \partial^2 Loss }{ \partial^2 \mu_k } &=  
 \frac{ \partial  \E_{\mu,\theta} ( D_k )  }{ \partial \mu_k } + 2\cdot\lambda_\mu \\
 &= n \cdot \frac{ \partial  \P_{\mu,\theta} ( D_k =1)  }{ \partial \mu_k } + 2\cdot\lambda_\mu, 
\end{align*}
($\mu_k$ is the $k^{th}$ entry of the parameter $\mu$)

\noindent
here, we may note that $ \P_{\mu,\theta}( D_k =1) $ is a logistic function of parameter $ \mu_k $:
Indeed, we note  $ a \equiv \exp(-\mu_k) \sum\limits_{x,y} \phi_k(x) \exp(\phi(x) \cdot \mu + y \phi(x) \cdot \theta ) $does not depend on $\mu_k$,  and neither does $b\equiv  \sum\limits_{x,y} (1-\phi_k(x)) exp(\phi(x) \cdot \mu + y \phi(x) \cdot \theta ) $
We then have:
\begin{align*}
     \P_{\mu,\theta} ( D_k =1)  &= \sum\limits_{x,y} \phi_k(x) \exp(\phi(x) \cdot \mu + y \phi(x) \cdot \theta ) \frac{1}{Z}  \\
     &= \exp(\mu_k)  \frac{a}{ \exp(\mu_k) \cdot a  + b  }.
\end{align*}

\noindent
Thus its derivative is: 
$$ \frac{ \partial  \P_{\mu,\theta} ( D_k =1)}{ \partial \mu_k  } =   \P_{\mu,\theta} ( D_k =1) \cdot (1-  \P_{\mu,\theta} ( D_k =1) ) $$ 

\noindent
And finally:
\begin{align*}
    \frac{ \partial^2 Loss }{ \partial^2 \mu_k } &=  
 \frac{ \partial  \E_{\mu,\theta} ( D_k )  }{ \partial \mu_k } + 2\cdot\lambda_\mu \\
 &= n \cdot  \P_{\mu,\theta} ( D_k =1) \cdot (1-  \P_{\mu,\theta} ( D_k =1) ) + 2\cdot\lambda_\mu  \\
 & \approx n \cdot  \P_{\mu,\theta} ( D_k =1)  + 2\cdot\lambda_\mu \\
  & \approx  \E_{\mu,\theta} ( D_k ) + 2\cdot\lambda_\mu 
\end{align*}

We used this formula, with the same estimator of $ \E_{\mu,\theta}(D)$ as the one already used for estimating the gradient from the Gibbs samples. The derivation of the diagonal Hessian on $\theta$ is perfectly similar.

\end{document}